\documentclass[letterpaper, 10pt, conference]{ieeeconf}  

\IEEEoverridecommandlockouts 

\usepackage[nolist,nohyperlinks]{acronym}
\usepackage{amsmath,amssymb,amsfonts}
\usepackage{booktabs}
\usepackage{hyperref}
\usepackage{graphicx}
\usepackage{overpic}
\usepackage{xcolor}
\usepackage{adjustbox}

\acrodef{PDF}[PDF]{probability density function}
\acrodef{KDE}[KDE]{kernel density estimator}
\acrodef{MLE}[MLE]{maximum likelihood estimation}
\acrodef{DOF}[DoF]{degree of freedom}
\acrodef{IMU}[IMU]{inertial measurement unit}
\acrodef{FT}[F/T]{force/torque}
\acrodef{GRU}[GRU]{gated recurrent unit}
\acrodef{MLP}[MLP]{multi-layer perceptron}
\acrodef{NMN}[NMN]{neural measurement network}

\def\bfm{\mathbf{m}}

\def\bfq{\mathbf{q}}

\def\bfu{\mathbf{u}}
\def\bfv{\mathbf{v}}

\def\bfgamma{\boldsymbol{\gamma}}
\def\bfomega{\boldsymbol{\omega}}

\title{
    A Data-driven Contact Estimation Method for Wheeled-Biped Robots
}

\author{U. Bora G\"{o}kbakan$^{1}$, Frederike D\"{u}mbgen$^{1}$ and St\'{e}phane Caron$^{1}$
\thanks{$^{1}$ The authors are with Inria and the Computer Science Department of ENS (DI ENS), PSL Research University, Paris, France. Corresponding author: \texttt{umit-bora.gokbakan@inria.fr}}}

\begin{document}

\maketitle
\thispagestyle{plain} 
\pagestyle{plain} 


\begin{abstract}
    Contact estimation is a key ability for limbed robots, where making and breaking contacts has a direct impact on state estimation and balance control. Existing approaches typically rely on gate-cycle priors or designated contact sensors. We design a contact estimator that is suitable for the emerging wheeled-biped robot types that do not have these features. 
    To this end, we propose a Bayes filter in which update steps are learned from real-robot torque measurements while prediction steps rely on inertial measurements. We evaluate this approach in extensive real-robot and simulation experiments. Our method achieves better performance while being considerably more sample efficient than a comparable deep-learning baseline.
\end{abstract}


\section{Introduction}

Legged robots with feet can navigate rough terrains or confined spaces, yet at a relatively higher cost of transport and mechanical complexity than their wheeled counterparts. At the intersection between legs and wheels, wheeled-legged robots~\cite{klemm_ascento_2019,bjelonic_wheeled_legged} combine legs for active suspension and locomotion over varied terrains. They can drive, walk, jump~\cite{klemm_ascento_2019} or skate~\cite{geilinger_skaterbots_2018, bjelonic_skating_2018}.

Wheeled-legged robots make and break contact with their environment, for instance, when jumping or stepping downstairs. In the resulting hybrid dynamics, the presence or absence of contacts changes the equations of motion, thus impacting state estimation and motion control. Hence, the question of \emph{contact estimation} or contact detection, where the system determines its contacts from sensory measurements, has been studied in itself as a key component of legged robotic systems.

One straightforward approach to contact estimation is to add dedicated contact sensors to the system. Direct measurements of contact forces by a collocated \ac{FT} sensor were spearheaded by the Honda P2~\cite{hirai1998development} and were kept in following iterations of the humanoid robotics project~\cite{goswami_hrp-4_2019}. This design choice is still actively studied today, with recent works including an unsupervised approach by fuzzy \textit{c}-means clustering~\cite{rotella_2018_contact_learning} and a supervised approach training a contact-detection \ac{MLP}~\cite{piperakis2022robust}.

Alternative direct contact sensors include load cells and additional \acp{IMU}. The Solo open-source quadruped robot, in its first iteration~\cite{grimminger2020open}, included a load cell in each foot, while measurements from collocated \acp{IMU} were considered in~\cite{maravgakis_probabilistic_2023}. Overall, the upsides of dedicated contact sensors include the reliability and simplicity of estimating contact from direct measurements. The downsides include increased leg inertia, design complexity, and hardware costs.

\begin{figure}[t]
\centering
\includegraphics[width=\linewidth]{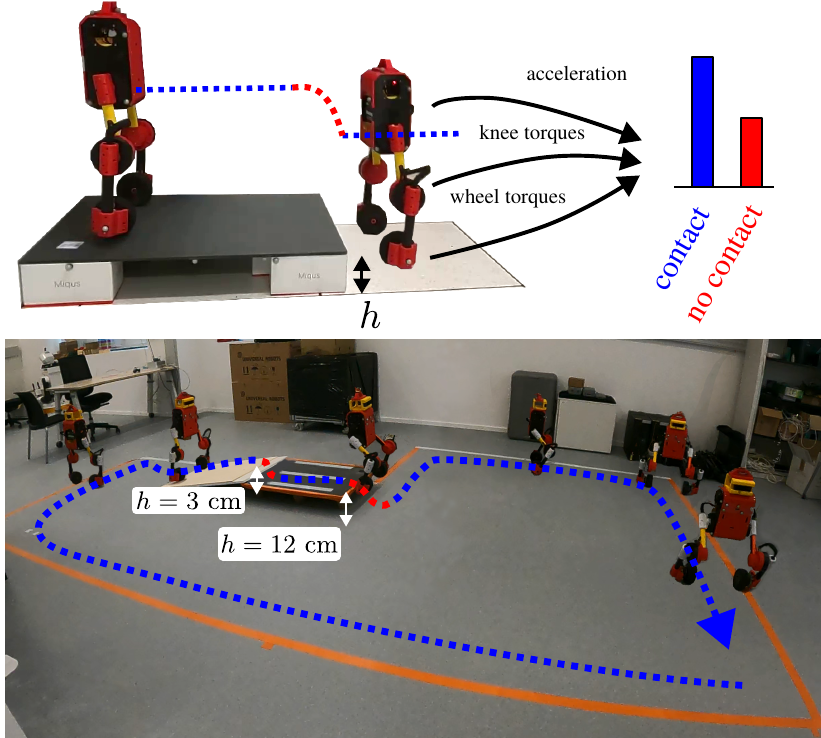}
\caption{Robustly detecting the moments when a wheeled-biped robot makes and breaks contact is crucial for successful estimation and control. This paper proposes a contact estimator based only on inertial and torque measurements. The measurements are fed into a novel Bayesian filter formulation to robustly estimate the binary contact state. We validate our results extensively both in simulation and real-world experiments, as depicted in the bottom figure.}
\label{fig:experiments}
\end{figure}

More recent humanoid~\cite{englsberger2014overview} and quadruped~\cite{hutter2012starleth,seok2013design,hutter_anymal_2016} robots explored a different design path, using integrated joint torque sensors rather than dedicated contact sensors. This \emph{indirect} approach made the question of \emph{contact estimation} more pressing, as fault-prone contact detection was identified as a major cause of drift by ensuing state estimators~\cite{Bloesch-RSS-12}. Following up on this observation, \cite{hwangbo_probabilistic_2016} proposed a probabilistic state machine to estimate each leg's contact state separately by fusing the system's kinematic and dynamic models. The approach was later extended in~\cite{jenelten_dynamic_2019} to distinguish between sticking and sliding contact modes. Nevertheless, these methods rely on gait priors and kinematic models, which are not informative for stiff-legged wheeled-bipeds.

In this work, we estimate the ground contact state of a wheeled biped robot using only \ac{IMU} and joint torque sensors that are readily available on current robots. Our main contributions are 1) the design of a Bayes filter with parameters learned from real-robot data, 2) its noise analysis in simulation and real-robot performance evaluation compared with a deep-learned baseline, and 3) the real-time, open-source implementation of the estimator, packaged as a new agent for the Upkie wheeled-biped platform. %

\begin{figure*}[ht]
    \centering
    \includegraphics[width=\linewidth]{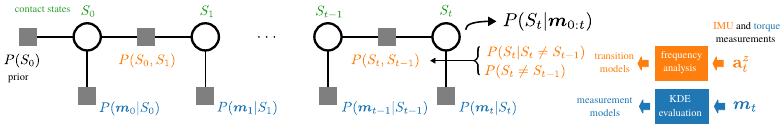}
    \caption{Overview of the estimation pipeline. We use Bayesian filtering to estimate the posterior probability $P(S_t|\mathbf{m}_{0:t})$ of being in contact state $S_t$ given measurements $\mathbf{m}_{0:t}$. Internally, measurement probabilities $P(\mathbf{m}_t | S_t)$ are estimated by kernel density estimation (KDE) from knee and wheel torque sensors, while transition probabilities are estimated from IMU measurements through frequency analysis.}
    \label{fig:graph}
\end{figure*}

\section{Related works}
\label{sec:relworks}

Our approach borrows the probabilistic state machine of~\cite{hwangbo_probabilistic_2016,jenelten_dynamic_2019}, which we cast into the framework of Bayesian filtering. Rather than using a kinodynamic model of the system, however, we rely directly on real-robot data to learn the underlying measurement model distributions and to derive transition probabilities.

In a more data-driven approach, \cite{camurri_probabilistic_2017} used logistic regression models to estimate the probability of ground contact, with parameters learned from real-robot data. This pipeline also relies on a kinodynamic model to compute ground reaction forces, while our proposal is to work directly with collected sensory data. In a more recent work,~\cite{maravgakis_probabilistic_2023} instrumented the feet of their robots with additional \ac{IMU} sensors, fitting a \ac{KDE} on each of the six \ac{IMU} measurements independently. Both~\cite{camurri_probabilistic_2017} and~\cite{maravgakis_probabilistic_2023} are purely measurement-based approaches, whereas in the probabilistic framework of Bayesian filtering we will consider two separate measurement and transition processes. We evaluate the effect of this separation as an ablation in our experiments and show that considering transition probabilities is important for high performance.

Training contact estimation from extensive simulation data has also been explored by~\cite{youm2024legged}, who suggested early stopping as an alternative to domain randomization for closing the sim-to-real gap more effectively. We reproduce the contact estimation network from their work and discuss it further in our experiments. We show that we achieve higher performance with significantly lower sample complexity by learning from a few real-world experiments. 

\section{Method}
\label{sec:methods}

We consider the question of contact estimation from a probabilistic standpoint. Our proposal is to model our system as a \emph{Markov process} and use Bayesian filtering to recursively update the contact estimate based on incoming measurements. Figure~\ref{fig:graph} gives an overview of the pipeline.%

\subsection{Probabilistic Framework}

We represent the contact state at time $t$ as a binary random variable $S_t$ that can take values $\{C, \lnot C\}$, where $C$ indicates contact with the ground and $\lnot C$ indicates no contact. We aim to calculate the posterior probability distribution of $S_t$, which we will refer to as the \emph{belief} following the convention of~\cite{thrun_probabilistic_2005}. It is defined as
\begin{equation}
    \text{bel}(S_t) := P(S_t \ | \ \mathbf{m}_{0:t}),
\end{equation}where $\mathbf{m}_{0:t}=\{\mathbf{m}_0, \cdots,\mathbf{m}_t\}$ contains all measurements made up to time $t$. The belief at time $t > 0$ can be recursively computed by alternating between a prediction and an update step, defined respectively by
\begin{subequations}
    \begin{align}\label{eq:predict}
         \overline{\text{bel}}(S_t) &= \sum_{S\in{C,\lnot C}} P(S_t\, |\,S_{t-1}=S)\text{bel}(S_{t-1}), \\
        \text{bel}(S_t) &= \eta_t^{-1} P(\mathbf{m}_t\,|\,S_t) \overline{\text{bel}}(S_t),\label{eq:update}
    \end{align}
\end{subequations}
where we have introduced the intermediate prior belief $\overline{\text{bel}}(S_t):=P(S_t\,|\,\mathbf{m}_{0:t-1})$, and the normalization constant $\eta_t$. The prediction step updates the current belief based on the \textit{transition model} $P(S_t\, |\, S_{t-1})$ (also called \textit{process} or \textit{motion} model), while the update step incorporates new observations through the \textit{measurement model} $P(\mathbf{m}_t\,|\,S_t)$. We discuss the form of these two models in the next sections. The normalization factor $\eta_t$ in  \eqref{eq:update} is estimated by marginalizing over both possible contact states:
\begin{equation}
    \eta_t = \sum_{S_t \in \{C, \lnot C\}} p(\mathbf{m}_t\,|\,S_t = S)\overline{\text{bel}}(S_{t}=S).
\label{eq:normalization_term}
\end{equation}

In the absence of other information, we define a flat (non-informative) prior $P(S_0=C) = P(S_0=\lnot C) = 0.5$ for the initial contact state.

\subsection{Measurement Model}

The measurement model $P(\mathbf{m}_t\,|\,S_t)$ is given by the likelihood of observing the measurements $\mathbf{m}_t$ given the contact state $S_t$. Rather than using a parametric model, such as a kinodynamic robot model, we follow a non-parametric approach by fitting a \ac{KDE} to torque measurements from the knee and wheel joints of each leg:
\begin{equation}
    \mathbf{m}_t = (\tau_{\text{knee}}, \tau_{\text{wheel}}) \in \mathbb{R}^2
\end{equation}

\acp{KDE} estimate the \ac{PDF} of a random variable by placing a kernel function $K$ over all data points and summing them up. The kernel function is a non-negative function with a bandwidth parameter $h$ that trades off smoothness and accuracy of the estimated \ac{PDF}, whose formula is given by:
\begin{equation}
    f_S(\mathbf{x}) = \frac{1}{N_S} \sum_{i=1}^{N_S} K\left(\mathbf{x} - \mathbf{\hat{x}}_i^S, h\right)
\end{equation}
where $K$ is the kernel function, $\mathbf{\hat{x}}_i^S$ are the data points collected for a given contact state $S$, and $N_S$ is the number of data points in the measurement set.

\begin{figure}
    \centering
    \vspace*{0.5cm}
    \includegraphics[width=\linewidth]{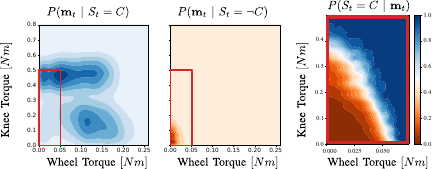}
    \caption{Results of modeling torque measurements $\mathbf{m}_t$ in contact ($C$) and no-contact ($\lnot C$) states. The left two plots show the Gaussian KDEs fit to vectors of absolute knee and wheel torques $\mathbf{m}_t \in \mathbb{R}^2 : (\tau_{\text{knee}}, \tau_{\text{wheel}})$. The contact likelihood model is normalized by the marginal likelihood by assuming a flat $\overline{\text{bel}}(S_t)$ at each timestep. This gives a simple non-recursive ``Measurement Only'' model to estimate the contact probability directly, without any dependence on the history or prior beliefs (right-most plot).}
    \label{fig:kdes}
\end{figure}

This \ac{PDF} provides our measurement model:
\begin{equation}
    P(\mathbf{m}_t\ | \ S_t = S) = f_{S}(\mathbf{m}_t),
\end{equation}
where $f_S$ is the KDE fitted on contact state $S$. As we model contact as a binary variable, we only have two \acp{KDE} $f_C$ and $f_{\lnot C}$ to fit. We use Gaussian kernels, as they are smooth functions with infinite support that allow us to cover the entire space of torque measurements.
Figure~\ref{fig:kdes} illustrates the KDEs estimated from real-robot data.

\subsection{Transition Model}

Given a pair $A, B \in \{C, \lnot C\}$, we estimate the transition probability $P(S_t = A \,| \,S_{t-1} = B)$ empirically from accelerometer measurements. It is possible to detect when a \emph{transition event} occurs thanks to nonstationarities in accelerometer measurements. It is nevertheless not straightforward to predict the \emph{direction} (landing or takeoff) of the transition, as empirically, both takeoff and landing yield high upward and downward accelerations indifferently. We approach this by decomposing transition probabilities into two factors:
\begin{enumerate}
    \item \emph{Switch} (non-directional) probabilities $P(S_t \neq S_{t-1})$ that treat takeoff and landing indifferently, and
    \item \emph{Directional} probabilities $P(S_t = C\,|\,S_t \neq S_{t-1})$ and $P(S_t = \lnot C\,|\,S_t \neq S_{t-1})$ that estimate the direction of a given switch.
\end{enumerate}
Transition probabilities are then a product of these two. For instance, in case of a landing, we have:
\begin{align}
    &P(S_t = C, S_{t-1} = \lnot C) \nonumber \\
    &= P(S_t = C, S_t \neq S_{t-1}) \\
    &= P(S_t = C\,|\,S_t \neq S_{t-1})\,P(S_t \neq S_{t-1}).
\end{align}

\subsubsection{Switch Probabilities}

We use the vertical accelerometer measurements for calculating switch probabilities as follows. Let us denote by $\bfgamma_t \in \mathbb{R}^3$ the accelerometer measurement from the \ac{IMU} at a given time step $t$. We consider a slice of the $N$ most recent accelerometer measurements projected on the vertical axis in the world frame:
\begin{equation}
    \mathbf{a}_t := \left[\bfgamma_{t-N+1}^z, \bfgamma_{t-N+2}^z, ..., \bfgamma^z_t\right] \in \mathbb{R}^N
\end{equation}
and compute its power spectrum $A_t(\omega)$:
\begin{equation}
    A_t(\omega) = \frac{\|{\hat{\mathbf{a}}_t(\omega)}\|^2_2}{N}
    \label{eq:power_spectrum}
\end{equation}
where $\hat{\mathbf{a}}_{t}(\omega)$ is obtained by performing the short-time Fourier transform (STFT) of the windowed signal, and $\omega$ is the frequency bin. 

Both take-off and landing events are impulsive, meaning their respective windows will have higher power densities than when neither occurs. We thus formulate the transition probability with a sigmoid function $\sigma_{1}$ of the power density in the window:%
\begin{equation}
    P(S_t \neq S_{t-1}) = \sigma_1(A_t(\omega)) %
    \label{eq:nondirectional_prob}
\end{equation}
where $\sigma_1$ is a sigmoid with offset and slope set to $(8, 1)$ in all experiments.

\subsubsection{Directional Transition Probabilities}

On real-robot data, we identified that vertical accelerations were higher during landing, by a significant margin, compared with take-off. These Dirac-like accelerations are reflected in a more uniform energy distribution across the frequency spectrum. Takeoffs, on the other hand, are smoother and display relatively more energy at lower frequencies. We then distinguish the two with a heuristic based on the median frequency $\omega_{\text{median}}$ splitting the energy mass of the sliding window into two halves\footnote{As we deal with discrete time and frequencies, we interpolate linearly between the two frequencies closest to the true median.}:
\begin{equation}
    \omega_{\text{median}}(A_t(\omega)) = \arg\min_{\omega^*} \int_{0}^{\omega^*} A_t(\omega)\,\mathrm{d}\omega - \int_{\omega^*}^{+\infty} A_t(\omega)\,\mathrm{d}\omega.%
\end{equation}

We convert $\omega_{\text{median}}$ to a directional probability using a logistic function  $\sigma_2$ with offset and slope given by $(7.5, 2.5)$: 
\begin{equation}
    P(S_t = C\,|\,S_t \neq S_{t-1}) = \sigma_{2}\left(\omega_{\text{median}}(A_t(\omega))\right).
\end{equation}

\begin{figure}[tb]
\centering
\vspace*{0.5cm}
\includegraphics[width=0.9\linewidth]{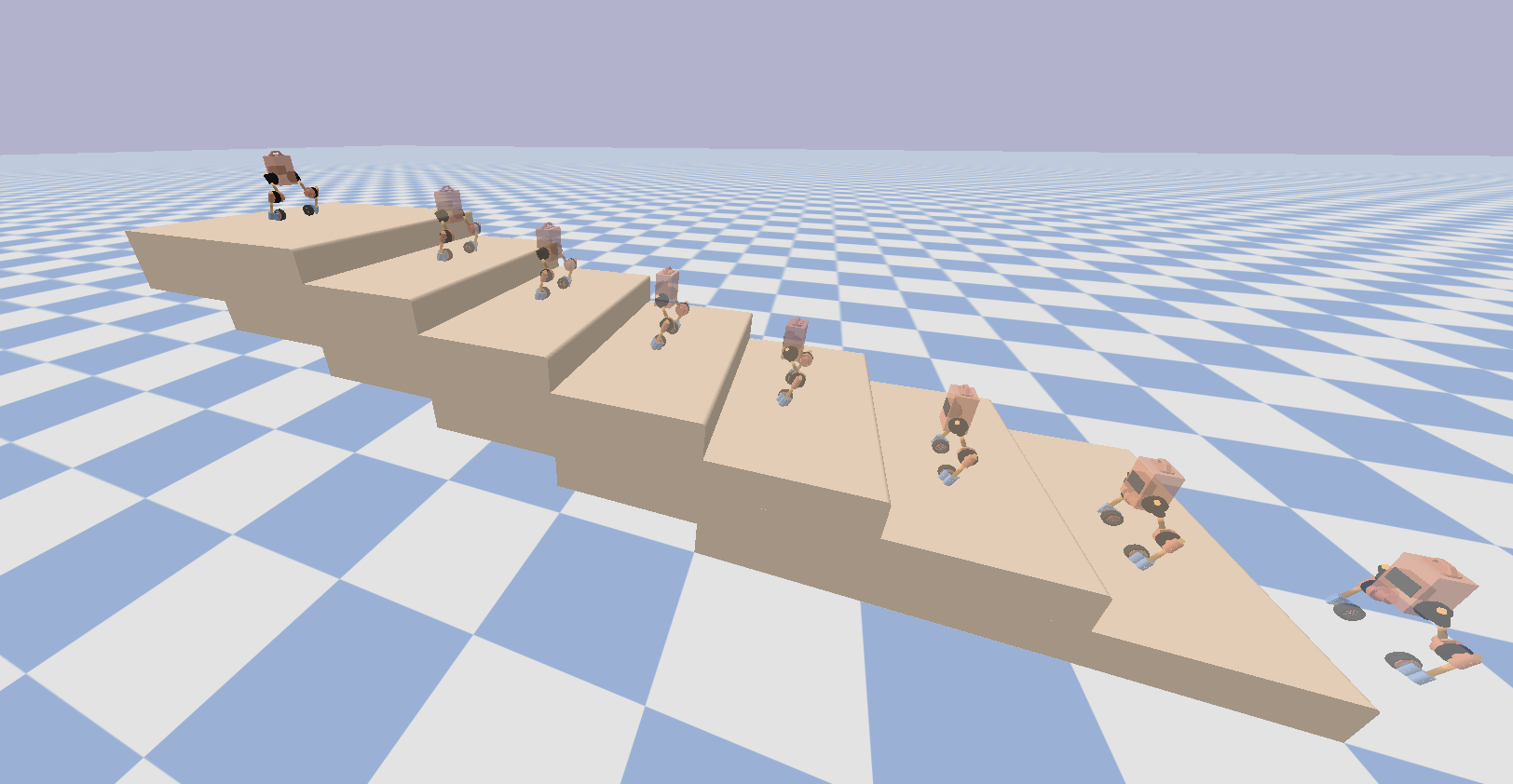}
\caption{The Bullet environment used to collect contact data. The robot was driven down a flight of steps, each with a height of 0.25. The captured IMU and proprioceptive readings were used to evaluate the contact estimator against noise.}
\label{fig:sim_stairs}
\end{figure}

\begin{figure*}[htb]
    \centering
    \vspace*{0.5cm}
    \includegraphics[width=.9\linewidth]{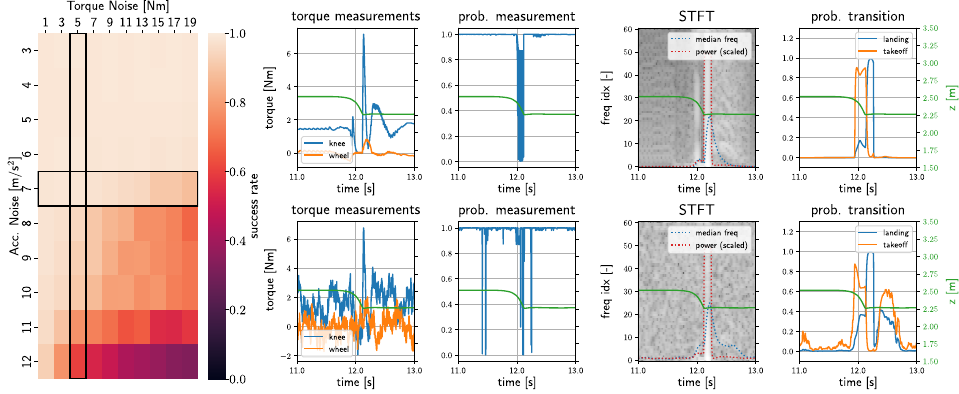}
    \caption{Robustness study with respect to measurement noise in simulation. The heatmap shows the overall success rate of the estimator under different noise levels (standard deviation of zero-mean Gaussian measurement noise). On its right, we provide a detailed study for two representative noise settings, highlighted with black rectangles. We zoom in on the robot rolling down one step (z-coordinate in world frame shown in green) and show, from left to right, the ground truth (top) and noisy (bottom) torque measurements, the resulting contact probabilities, the STFT of the ground truth (top) and noisy (bottom) accelerometer data (including power and median frequency), and the resulting transition probabilities.}
    \label{fig:sim_results}
\end{figure*}

\section{Simulation study}
\label{sec:simulation}

We study the sensitivity of the designed filter to sensory noise in simulation. The simulation environment, shown in Figure~\ref{fig:sim_stairs}, is built on \textit{Bullet} and publicly available. In the environment, the open source Upkie wheeled biped robot~\cite{upkie} is driven down a flight of steps of height 0.25~m each. We collect two datasets: one to fit the measurement model (one run with no additional noise) and one with many runs to test the estimator's performance by adding zero-mean Gaussian noise of varying magnitudes to knee torques, wheel torques, and accelerometer measurements. 

Using the ground-truth contact state, we measure the success rate as the number of correctly classified contact states, where we use a threshold of 0.8 on the probabilities to define contact. This resulting success rate is visualized as a function of the noise levels in Figure~\ref{fig:sim_results}. Across all acceleration noise levels, we observe a near-perfect success rate if the noise on torques is below 1~Nm. This is far above the noise levels we observed in real-world experiments (up to 0.12~Nm for wheel torques and 0.25~Nm for knee torques). The performance with respect to the acceleration noise is also near-perfect up to noise levels of about 7~m/s$^2$ , and real-world data was always below 0.23~m/s$^2$. 

We finally provide qualitative plots of the measurement and transition model performance to investigate how the different elements affect the drop in performance as we increase the noise. At the considered torque noise level, we observe many false non-contact detections, however this affects performance only little, possibly because the transition probability smoothes out such false detection. For the transition model, the added noise particularly affects the median frequency, which detects contact switch directions. This sensitivity leads to a drastic performance drop as we increase the noise (see, for example, the acceleration noise 6 in the the left heatmap in Figure~\ref{fig:sim_results}).

\section{Experiments}
\label{sec:results}

We implement and test the performance of the Bayesian filter to detect contacts on the wheeled-biped Upkie. Our study focuses on drops from unforeseen steps, where the robot alternates between contact and free-fall phases. We set up a test course where the robot travels over a horizontal floor, climbs a meter-long ramp with an 8\textdegree{} incline, drops 3 cm from the ramp onto a platform, rolls forward and drops again from a 12-cm step, which we repeated for a total number of 10 drops or 5 laps.\footnote{Falls took slightly longer than a true free-fall because of the rolling down phase.}

\subsection{Data collection}
We fit measurement models and transition functions using real-robot data collected during a 20-minute session. For measurement likelihoods, we collected data for 7 min where the robot either (i) balanced and moved around the room, or (ii) was hung on a tether without making contact with the ground. Figure~\ref{fig:kdes} illustrates the resulting distributions of torque measurements for both contact states.

In the second 13-minute long part of the data collection, we put the robot on an elevated platform (Fig. \ref{fig:experiments}, top) and drove it over the edge to induce a drop. After each drop, we wait for the robot the stabilize (helping it with a tether if needed), and reposition it on the platform. We use robot measurements and a relatively high-speed camera to segment each trial into three consecutive phases: \emph{rolling}, \emph{falling}, and \emph{stabilization}. We define \emph{takeoff} and \emph{landing} as the start and end times of each \emph{falling} phase.

\subsection{Training}

\paragraph{Measurement Model} As we collected \emph{contact} and \emph{no-contact} joint measurements separately, we can use them directly without annotation to fit our two \acp{KDE}. We estimate the bandwidth parameter $h$ ($\approx 12.5$) of our kernel functions following Scott's rule of thumb, based on the number of samples and dimensions~\cite{ScottBandwith}.

As the \ac{KDE} method has a time and memory complexity of $\mathcal{O}(nd)$ in the number of samples $n$ and dimensions $d$, it is common to use efficient search algorithms like balltrees or k-d trees and sample only near the input point\footnote{This is an approximation when the kernel has infinite support.}. In our case, we build $200\times200$ regular lookup grids and interpolate them linearly to achieve even faster prediction times, which is possible as our data has low-enough dimensionality.

\paragraph{Ablation} We integrate KDEs to predict the measurement likelihoods $P(\mathbf{m}_t\ | \ S_t)$ in the overall filter, but it would also be possible to normalize them to estimate a contact probability directly as: 
\begin{equation}
    P_{\tau}(S_t \ | \ \mathbf{m}_t ) = \frac{P(\mathbf{m}_t\ | \ S_t = C)}{\sum_{S \in \{C, \lnot C\}} P(\mathbf{m}_t\ | \ S_t = S)},
\end{equation}
We will call this indicator ``measurements only'', as it is an ablation of the Bayes filter without transitions.

\subsection{Neural Measurement Network}

We reproduce the proposal from \cite{youm2024legged} of a \emph{neural measurement network} (NMN) trained from simulated data produced by running an independent locomotion policy. Practical benefits from training in simulation include the ability to work with a large dataset and compare against perfect ground truth, the downside being a corresponding sim-to-real gap. While the adaption from point-foot quadruped to wheeled biped requires some changes, we strive to follow the original proposal whenever possible.

\paragraph{NMN inputs} In the formulation of~\cite{youm2024legged} for quadrupeds, the NMN receives as input a measurement vector $\bfm_t = [\bfgamma_t, \bfomega_t, \bfq_t, \dot{\bfq}_t, \bfu_{\textit{quad},t-1}]$, where $\bfgamma_t$ denotes the linear acceleration measured by the accelerometer of the \ac{IMU}, $\bfomega_t$ is the body angular velocity from the \ac{IMU} frame, $\bfq_t$ is the vector of measured joint angles with $\dot{\bfq}_t$ its time-derivative, and $\bfu_{\mathit{quad},t-1} = \bfq_{\mathit{des},t-1}$ is the vector of joint position targets sent at the previous time step to actuators. Rather than a position-controlled quadruped robot, in these experiments, our system is a wheeled biped with locked hip and knee positions and velocity-controlled wheels. We thus adapted the input measurement vector to:
\begin{equation}
    \bfm_t = [\bfgamma_t, \bfomega_t, \bfq_t, \dot{\bfq}_t, \bfu_{\mathit{wheel},t-1}]
\end{equation}
where \ac{IMU} and joint measurements have the same definition as above, and $\bfu_{\mathit{wheel},t-1}$ is the commanded ground velocity at the previous time step. In both cases, $\bfu_{\mathit{quad},t-1}$ and $\bfu_{\mathit{wheel},t-1}$ represent the action sent by the agent to its environment.

\paragraph{NMN outputs} The NMN outputs (1) a probability $P_{\textsc{NMN}}(S_t = C | \bfm_t)$ that the robot's feet are in contact with the ground, and (2) an estimate of the body linear velocity ${}_b \bfv_t(\bfm_t)$ from the \ac{IMU} frame to the world frame. Although we do not use the latter thereafter, we included it in our reproduction as it can be beneficial for the estimator to be trained to predict both outputs rather than training to detect contacts only.

\paragraph{Simulation} We train the \ac{NMN} in the open-source Bullet physics simulator~\cite{bullet}. The terrain consists of a flat horizontal floor, with a lateral friction coefficient of $\mu = 0.1$ and a rolling friction coefficient of $\mu_{R} = 0.01$. As recommended for wheels in the simulator documentation~\cite{bullet}, we set the ``stiffness'' and ``damping'' parameters (of the underlying frictional-contact linear complementarity problem) to $3 \cdot 10^4$ and $10^3$ respectively. During each episode, we randomly apply vertical external forces to lift the robot up or let it fall down, generating a variety of \emph{contact} and \emph{no contact} states.

\paragraph{Training} The network architecture from~\cite{youm2024legged} consists of a one-layer \ac{GRU} with $128$ hidden features, followed by a $256 \times 128$ \ac{MLP}. We generate a dataset of 4,400,000 simulation steps ($\approx$ 5.5 hours of simulation data), each step lasting $5~\mathrm{ms}$, then split it into a training set of 4,000,000 steps and a validation set of 400,000 steps. As in~\cite{youm2024legged}, at each iteration of the \ac{NMN} training loop, we use a batch size of 400 parallel environments rolling out episodes of 400 simulation steps using an independent locomotion policy. We select a sequence length of 10 to train the recurrent layer, trading off between the value of 1 used at inference (faster training, higher validation loss) and larger values such as 100 (slower training, lower validation loss). We use the Adam optimizer~\cite{kingma2014adam} with a learning rate of $5 \cdot 10^{-4}$ and 32 epochs.

As advocated in~\cite{youm2024legged}, we rely on early stopping rather than domain randomization over geometric and inertial parameters. We tracked training and validation losses to determine the number of iterations to stop at, which we identified as 20 in our use case, as opposed to 200 for a quadruped walking over varied terrains. This difference in scale can be readily explained by the lower number of degrees of freedom and terrain complexity of our application.

\subsection{Results}

We evaluate the complete Bayes filter, the simulation-trained NMN and the measurements-only ablation on the final task of detecting contacts on the real robot.

\paragraph{Contact estimation} We validate contact estimation as a binary classification task over a test set of real-robot data manually annotated by a human operator. We threshold the scalar output of each method to obtain a corresponding binary classification variable $\hat{\mathbf{y}}_t \in \{C, \lnot C\}$. We then evaluate ``point-wise'' \emph{precision} and \emph{recall} for each, as reported in Table~\ref{table:classification_results}.

All methods achieve (near-)perfect results when the robot is in contact, as the robot stays in contact for longer periods, where delays in transitions and label noise have very little effect. Nevertheless, predictions made during the free-flight phase are much more sensitive to errors and delays, as they are much shorter in duration. Despite this, we observe that both the Bayes filter and the measurement-only method maintain their ability to detect flight phases while predicting contact. 

\begin{table}[!htbp]
\centering
\begin{tabular}{c|*6c}
\toprule
 &  \multicolumn{2}{c}{$S_t = C$} & \multicolumn{2}{c}{$S_t = \lnot C$}\\
\midrule
{Method} & Precision & Recall & Precision & Recall \\
\midrule
\textbf{Bayes filter} & \textbf{1.0} & \textbf{1.0} & \textbf{0.55} & 0.37 \\
Meas. only & \textbf{1.0} & \textbf{1.0} & 0.53 & \textbf{0.40} \\
NMN \cite{youm2024legged} & 0.99 & \textbf{1.0} & 0.28 & 0.15 \\
\bottomrule
\end{tabular}

\vspace{0.5em}
\caption{Precision and recall of point-wise contact-state detection.}
\label{table:classification_results}
\end{table}

\paragraph{Transition Detection} Evaluating contact detection as a classification problem does not take into account the time-series nature of the collected data. In particular, delays in takeoff detection have a considerable impact on no-contact \emph{recall} and \emph{precision}, as delays account for a significant portion of the short flight duration. Consequently, we also evaluate the (mis)-detection rate of each method for landings and takeoffs. Results are reported in Table \ref{table:transition_detection}. After each transition, we employ a period of 250~ms to exclude irrelevant (mis)detections.

\begin{table}[!htbp]
\centering
\begin{tabular}{c|*4c}
\toprule
Method &  \multicolumn{2}{c}{Takeoff} & \multicolumn{2}{c}{Landing}\\
\midrule
{}  & Precision & Recall & Precision & Recall \\
\textbf{Bayes filter} & \textbf{0.91} & \textbf{1.0} & \textbf{0.71} & \textbf{1.0} \\
Meas. only & 0.59 & \textbf{1.0} & 0.55 & \textbf{1.0} \\
NMN~\cite{youm2024legged} & 0.53 & 0.90 & 0.56 & 0.90 \\
\bottomrule
\end{tabular}
\vspace{0.5em}
\caption{Detection of transition events over 10 drops during a session of 4 minutes.}
\label{table:transition_detection}
\end{table}

Both the measurement ablation and the Bayes filter achieve perfect recall on both transition types. In other words, they never fail to detect a transition within the admitted period, while the NMN baseline misses one. It is also important to consider the precision of the transitions, as misdetections can potentially lead to inappropriate actions by the robot. The Bayes method achieves a considerably higher precision rate, meaning we can ascribe more confidence to its predictions.

We witness that using measurements only can lead to more frequent misdetections. A transition can be triggered due to a spontaneous change in leg torques, which is frequent for a wheeled bicycle during balancing or crouching.

\begin{table}[!htbp]
\vspace{0.25cm}
\centering
\begin{tabular}{c | c c}
\toprule
Method & Takeoff & Landing \\
\midrule
\textbf{Bayes filter} & \textbf{77.1 $\pm$ 31.8}  & 17.9 $\pm$ 9.1 \\
Meas. only & 83.1 $\pm$ 31.9 & \textbf{10.1 $\pm$ 7.37} \\
NMN~\cite{youm2024legged} & 125.0 $\pm$ 59.6 & 22.4 $\pm$ 7.4 \\
\bottomrule
\end{tabular}
\vspace{0.5em}
\caption{Transition latencies of tested methods (ms). Only correctly identified transitions were considered.}
\label{table:latencies}
\end{table}

In Table \ref{table:latencies}, the average time each method takes to detect a contact transition event is given. It is shown that, despite being a more conservative estimator that is not triggered as often as the measurement-only baseline, the Bayes method achieves favorable latencies.

\section{Conclusion and discussion}
\label{sec:conclusion}

We have presented a Bayesian filter for contact estimation on wheeled-biped robots, where all measurements come from IMU and joint torque sensors that are commonly available on experimental platforms. We learn measurement likelihoods from a small real-robot dataset of labeled torque measurements, and estimate transition probabilities online using accelerometer data. We have evaluated this contact estimator in real-robot experiments, where we observed better accuracy while requiring considerably fewer samples than an existing alternative trained on extensive simulation data.

The proposed method can be improved in several ways. First, because of the Markov assumption, the filter ignores longer-term effects, such as the low-frequency acceleration oscillations that can follow a harsh landing. Augmenting the state or using transition models with memory capabilities, such as long-short-term memory, could be a way to capture longer contexts. On another front, our method requires retraining for different types of robots. A future study could investigate the variability of learned distributions across different robots.


\section*{Acknowledgements}

This work was supported by the European Union through the AGIMUS project (GA No. 101070165), the PEPR O2R AS2 (No. ANR-22-EXOD-0006), and the ANR JCJC project NIMBLE (No. ANR-22-CE33-0008).

\addtolength{\textheight}{-10.8cm}   

\bibliographystyle{IEEEtran}
\bibliography{references}

\end{document}